\documentclass{svproc}

\usepackage[hidelinks]{hyperref} %
\usepackage{url} %
\usepackage{cite}
\usepackage{booktabs} %
\usepackage{amsfonts} %
\usepackage{amsmath}
\usepackage{amssymb}
\usepackage{nicefrac} %
\usepackage{microtype} %
\usepackage{doi}
\usepackage[caption=false, font=footnotesize]{subfig}
\usepackage[dvipsnames, table]{xcolor} %
\usepackage{multirow}
\usepackage{floatrow}
\usepackage{blindtext}
\usepackage{physics} %
\usepackage{bm} %
\usepackage{upgreek} %
\usepackage{listings} %
\usepackage[inline]{enumitem} %
\setlist{nosep} %
\usepackage{dblfloatfix} %
\usepackage{tikz}
\usepackage{xcolor}
\usepackage{balance}
\usepackage[edges]{forest} %
\usepackage[final]{showlabels} %
\usepackage{lipsum} %

\usepackage{mymacros}

\begin{document}

\title{Effect of Optimizer, Initializer, and Architecture of Hypernetworks on \\Continual Learning from Demonstration}
\titlerunning{Effect of Optimizer, Initializer, \& Architecture of Hypernets on CLfD}

\author{Sayantan Auddy\inst{1} \and Sebastian Bergner\inst{1} \and Justus Piater\inst{1,2}}
\authorrunning{Sayantan Auddy et al.}
\institute{
	Department of Computer Science, University of Innsbruck, Innsbruck, Austria \\
    \email{\{sayantan.auddy, justus.piater\}@uibk.ac.at} \email{sebastian.bergner@student.uibk.ac.at}
    \and
    Digital Science Center, University of Innsbruck, Innsbruck, Austria \\
}

\maketitle

\begin{abstract}
    In \emph{continual learning from demonstration} (CLfD), a robot learns a sequence of real-world motion skills \emph{continually} from human demonstrations. Recently, hypernetworks have been successful in solving this problem. In this paper, we perform an exploratory study of the effects of different optimizers, initializers, and network architectures on the continual learning performance of hypernetworks for CLfD. Our results show that adaptive learning rate optimizers work well, but initializers specially designed for hypernetworks offer no advantages for CLfD. We also show that hypernetworks that are capable of \emph{stable} trajectory predictions are robust to different network architectures. Our open-source code is available at \codeurl{}.
    
    
\end{abstract}

\section{Introduction}
\label{sec:intro}
    
Learning from Demonstrations (LfD)~\cite{billard2016learning} is an intuitive way for humans to train robots without explicit programming. While the majority of research on LfD addresses single skill acquisition, some recent methods investigate continual learning from demonstration (CLfD)~\cite{auddy_node, auddy_snode}, i.e. learning multiple LfD motion skills sequentially in an open-ended way. Auddy et al.~\cite{auddy_node} propose a system of hypernetwork-generated neural ordinary differential equation solvers (NODEs) for continually learning a sequence of real-world 6-DoF trajectory learning tasks (e.g.opening a box, shelving a bottle, pouring from a cup, etc.) from human demonstrations. 
More recent work~\cite{auddy_snode}, shows that by enforcing stable trajectory predictions through hypernetwork-generated \emph{stable} NODEs~\cite{kolter2019learning}, the continual learning performance is greatly enhanced in addition to the expected guarantee of non-divergent and safe trajectory predictions.
Hypernetworks have also been utilized for continual reinforcement learning with robots~\cite{huang2021continual,schopf2022hypernetwork}.
The popularity of hypernetworks for robotic continual learning is mainly due to desirable features such as not having to store and retrain on data of past tasks, negligible parameter growth with additional tasks, and low catastrophic forgetting~\cite{parisi2019CL_survey}.

In any deep learning system, many decisions related to the architecture and training need to be taken. 
In previous works on CLfD~\cite{auddy_node, auddy_snode}, the effect of different deep learning components on the continual learning performance of hypernetworks remains unexplored. These past works have either followed accepted best practices (e.g. Adam is the optimizer) or followed the defaults from prior work (e.g. hypernetwork architecture). 
In this paper, we conduct an exploratory study in which we evaluate the effect of three key deep learning factors on the performance of hypernetworks for CLfD:
\begin{enumerate*}[label=(\roman*)]
    \item \emph{optimization algorithms},
    \item \emph{initialization schemes}, and
    \item \emph{hypernetwork and target network architectures}.
    \end{enumerate*}
We adopt the RoboTasks9 dataset of real-world LfD tasks~\cite{auddy_snode} as a benchmark, and train hypernetworks and chunked hypernetworks continually on the 9 tasks of this dataset. Additionally, we evaluate two kinds of target networks (generated by the hypernetworks): NODE~\cite{auddy_node}, and stable NODE (\snode{})~\cite{auddy_snode}. 

Our results show that adaptive learning rate optimizers exhibit the best empirical performance, but an initializer designed for hypernetworks (Principled Weight Initialization~\cite{ChangFL20WeightInit}) does not outperform a good default choice (Kaiming~\cite{he2015delving}) for CLfD. We also show that when stable NODEs (\snode{}s) are used as the target network (i.e. the LfD trajectory predictions are non-divergent), the continual learning performance is mostly independent of the network architecture.

\section{Background}
\label{sec:back}

In this section, we provide a brief overview of the deep learning topics involved in this study and refer the reader to references cited in the text for further details.
\smallskip

\par\noindent \textbf{Continual Learning from Demonstration}: Learning from Demonstration (LfD)~\cite{billard2016learning} is a robot training paradigm where a robot learns motion skills from human demonstrations. A popular way of performing LfD is via kinesthetic teaching where a human physically guides a robot and shows it how to perform a particular motion task. The trajectories demonstrated by the human are recorded and used to learn a vector field~\cite{urain2020imitationflow,kolter2019learning} which can then be used by the robot to perform a similar motion as the demonstration (while also being robust to changes to the initial state). 
While typical LfD approaches focus on learning a single motion skill, the objective of \emph{continual} LfD is to learn and remember a sequence of different motion skills, one at a time, in an open-ended manner with a single model and without storing training data of past demonstrations. In the past, this has been achieved by generating parameterized dynamical systems called \emph{Neural Ordinary Differential Equation solvers} (NODE)~\cite{chen2018neural} with Hypernetworks~\cite{auddy_node}. 
More recently, it has been shown that hypernetwork-generated \emph{stable} NODEs (NODEs augmented with a stabilizing Lyapunov function)~\cite{kolter2019learning} produce stable, non-divergent trajectories and are more effective at continually learning sequences of real-world and high-dimensional LfD tasks~\cite{auddy_snode}.
\smallskip

\par\noindent \textbf{Hypernetworks}: A hypernetwork is a meta-model that generates the parameters for another neural network called the \emph{target network}~\cite{von2019continual}. A hypernetwork $\mathbf{f}$ with parameters $\mathbf{h}$, takes as input a trainable task embedding vector $\mathbf{e}^m$ and generates the target network parameters $\mathbf{f}(\mathbf{e}^m,\mathbf{h}) = \uptheta^m$ for the $m  ^\upth$ task. A two-stage optimization process is employed to optimize $\mathbf{h}$ and the task embedding vector $\mathbf{e}^{m}$. In the first step, a candidate change $\Delta \mathbf{h}$ is computed by minimizing only the task-specific loss $\mathcal{L}^m$ of the current $m^\upth$ task. In the second stage, the $\Delta \mathbf{h}$ computed in the first step is used for optimizing the hypernetwork by minimizing the regularized loss 
$
    \hat{\mathcal{L}}^m = \mathcal{L}^m + \frac{\beta}{m-1} \sum_{l=0}^{m-1} ||\mathbf{f}(\mathbf{e}^l, \mathbf{h}^*) - \mathbf{f}(\mathbf{e}^l, \mathbf{h}+\Delta\mathbf{h})||^2
$, where $\beta$ is the regularization constant and $\mathbf{h}^*$ is the set of hypernetwork parameters before learning the $m^\upth$ task. Once the $m^\upth$ task is learned, the task embedding $\mathbf{e}^m$ is frozen and stored. For learning the $m+1^\upth$ task, a new task embedding $\mathbf{e}^{m+1}$ is initialized and the same two-step learning process is repeated.
A regular hypernetwork generates all the parameters $\uptheta^m$ of the target network from the final layer, which can result in a large parameter size. Alternatively, \emph{chunked} hypernetworks~\cite{von2019continual} generate the target network parameters in smaller segments called chunks, and consequently have a smaller size. See~\cite{von2019continual, auddy_node, auddy_snode} for details.
\smallskip

\par\noindent \textbf{Optimization Algorithms}: Stochastic Gradient Descent (SGD) updates parameters using gradients of the loss computed on randomly drawn mini-batches of the training data. SGD employs a fixed learning rate for all parameters, which can lead to sub-optimal convergence.
Root Mean Square Propagation (RMSProp)~\cite{tieleman20126} offers some advantages over SGD. It dynamically adapts learning rates for each parameter individually by normalizing the learning rates by an exponentially decaying average of the squared gradients. 
Adam~\cite{kingma2014adam} combines the advantages of AdaGrad~\cite{duchi2011adaptive} and RMSProp~\cite{tieleman20126}. Adam is invariant to the rescaling of gradients and is robust against sparse gradients. Adam also applies bias corrections for the first and second moments of the gradients. Although the choice of the optimizer is problem-dependent without a universally best option~\cite{goodfellow2016deep}, adaptive algorithms such as Adam are widely used due to their robustness.
\smallskip

\par\noindent \textbf{Initialization Schemes}: Initializers play a key role in neural network training. The Xavier/Glorot initializer~\cite{glorot2010understanding} is designed for the hyperbolic tangent activation and addresses the vanishing/exploding gradient problem by providing a suitable scale for weights. The Kaiming/He initializer~\cite{he2015delving} is designed for ReLU and its variants. It takes into account the asymmetric nature of ReLU and maintains a consistent variance in both the forward and backward passes of the network. Principled Weight Initialization for Hypernetworks (PWI)~\cite{ChangFL20WeightInit} initializes hypernetworks such that parameters of the generated target networks are scaled properly, to avoid vanishing/exploding gradients during hypernetwork training. Note that when conventional initializers (Xavier, Kaiming, etc.) are used in hypernetworks, the scale of target network parameters is not controlled.

\section{Experiments and Results}
\label{sec:experiments}

We train hypernetworks (HN) and chunked hypernetworks (CHN), each with either a NODE or an \snode{} as the target network, resulting in 4 kinds of hypernetworks (\hnnode{}, \chnnode{}, \hnsnode{}, \chnsnode{}).
We compare the performance of 3 different optimizers (SGD, RMSProp, Adam) and 3 different initializers (Kaiming, PWI, Xavier) when used to train the 4 kinds of hypernetworks. Furthermore, we evaluate 16 different architectures for each hypernetwork.
Each model is trained continually on the 9 LfD tasks of RoboTasks9~\cite{auddy_snode}. We report the widely used \emph{Dynamic Time Warping} (DTW) error~\cite{auddy_node, auddy_snode} that measures the difference between trajectory predictions and ground truth demonstrations.
Due to the large number of possible combinations, we perform our experiments in 3 stages to keep the number of runs manageable. To aid reproducibility and further research, our code and experiment hyperparameters are available at \codeurl{}.
\smallskip

\par
\begin{figure}[b]
    \includegraphics[width=\textwidth]{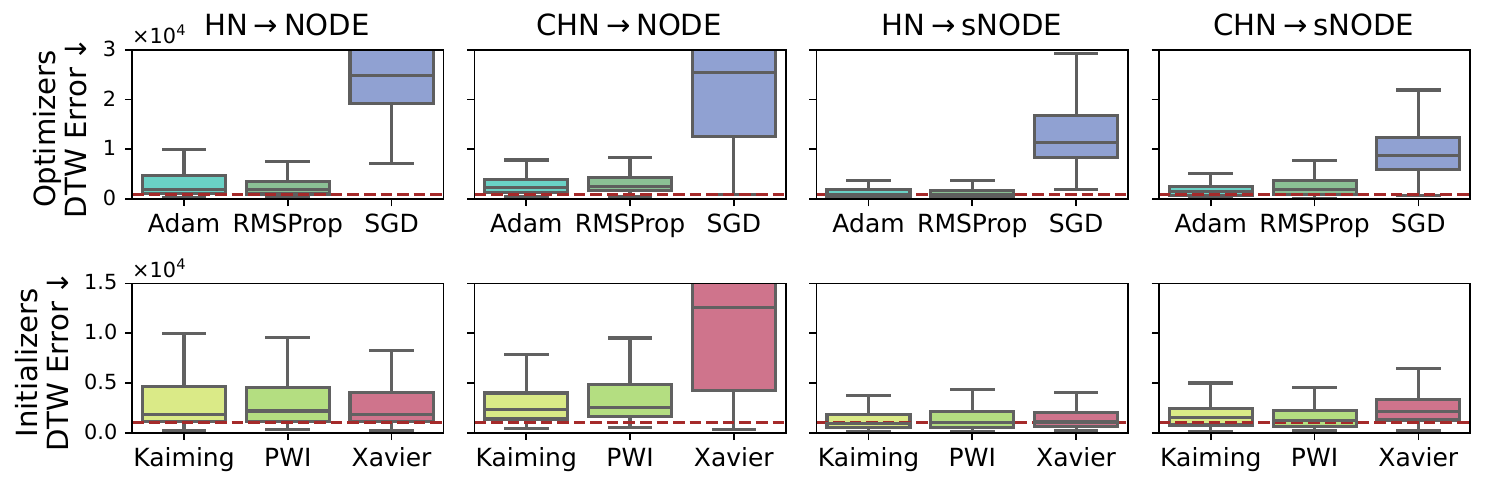}
    \caption{DTW errors (lower is better) of different \emph{optimizers} (top), and \emph{initializers} (bottom) for 5 independent runs. For reference, the dotted brown line shows the best possible median DTW score from~\cite{auddy_snode} (when each task is learned with a separate model).}
    \label{fig:exp_1_2}
\end{figure}

\noindent \textbf{Experiment 1 (Optimizers)}: We train each of the 4 kinds of hypernetworks (\hnnode{}, \chnnode{}, \hnsnode{}, \chnsnode{}) with the 3 different optimization algorithms (SGD, RMSProp, Adam). We use a fixed architecture (same as~\cite{auddy_snode}) and initializer (Kaiming) for all hypernetworks. After each task is learned during the continual learning process, we evaluate each model on the currently learned task and all previous tasks and repeat each run 5 times with independent seeds. \myfigure{fig:exp_1_2}~(top row) shows the overall DTW errors during this evaluation.
For all optimizers, \hnsnode{} and \chnsnode{} outperform \hnnode{} and \chnnode{}. The overall performance of SGD is much worse than Adam and RMSProp, both of which achieve similarly good DTW errors.
\smallskip

\par\noindent \textbf{Experiment 2 (Initializers)}: We use Adam as the optimizer since it is slightly better than RMSProp for the well-performing models (with \snode{}) in the previous experiment. We compare 3 initializers (Kaiming, PWI, and Xavier) while training the 4 kinds of hypernetworks with fixed architectures~(same as~\cite{auddy_snode}). We follow the same training and evaluation steps as in experiment 1 and repeat each run 5 times. We report the DTW errors for these evaluations in \myfigure{fig:exp_1_2} (bottom row).
All initializers perform similarly, except for \chnnode{}, where the Xavier initialization fails completely (very high DTW errors). For the other hypernetworks, all the initializers achieve similar results. All our hypernetworks use ReLU activations, and while Xavier is designed for $tanh/sigmoid$, we still include it in our experiments since it is used in a similar comparison in a prior work on hypernetwork initialization~\cite{ChangFL20WeightInit}. Additionally, though PWI is specially designed for hypernetworks, it does not outperform Kaiming for CLfD.
\smallskip

\par
\begin{figure}[t]
    \includegraphics[width=\textwidth]{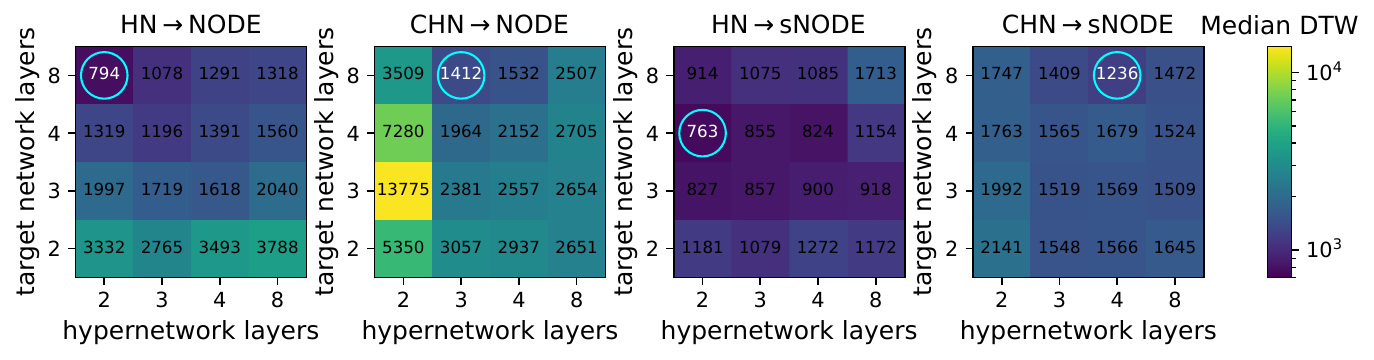}
    \caption{Effect of hypernetwork depth (y-axis) and target network depth (x-axis) on continual learning from demonstration. Each heatmap corresponds to a different hypernetwork type. Circled numbers show the best DTW for each hypernetwork. Colors are scaled logarithmically. Median values over 5 independent runs are shown.}
    \label{fig:arch}
\end{figure}

\noindent \textbf{Experiment 3 (Architecture)}: In our final experiment, we fix Adam as the optimizer and Kaiming as the initializer, as they achieve marginally better median DTW scores than their respective alternatives. In this experiment, we evaluate 4 different network depths (2, 3, 4, or 8 layers) for both the hypernetwork and its generated target network, resulting in 16 different architectures for each of the 4 kinds of hypernetworks. We also modify the number of units in each layer such that the overall parameter size of the networks is roughly similar and comparable to the network sizes of the previous experiments. We follow the same training procedure as the previous two experiments and repeat each run 5 times with independent seeds. The median DTW results are shown in \myfigure{fig:arch}. 

For \hnnode{}, the depth of the target network affects the overall performance much more than the depth of the hypernetwork, while for \hnsnode{}, almost all architectures achieve similar results irrespective of the depth of either network. Chunked hypernetworks (\chnnode{} and \chnsnode{}) on the other hand, perform best when the hypernetwork is 3-4 layers deep and the target network is 4-8 layers deep. However, similar to \hnsnode{}, \chnsnode{} also performs similarly for almost all architectures. In summary, hypernetworks with \snode{} as the target network perform well and are much less sensitive to the network architecture than hypernetworks with NODE that are not guaranteed to predict non-divergent trajectories.

\section{Conclusion}
\label{sec:conclusion}

We empirically demonstrated the effects of the optimizer, initializer, and network architecture on the performance of hypernetworks for continual learning from demonstration. Our findings show that adaptive learning rate optimizers (Adam, RMSProp) are a good choice for CLfD. As an initializer, Kaiming is a good default choice that performs as well as the PWI initializer which is specially designed for hypernetworks. We also showed that hypernetworks with \snode{} as the target network produce uniformly good results that are mostly independent of the network architecture. Our findings can help in making informed design decisions while developing hypernetwork-based methods for CLfD in the future.

\bibliographystyle{IEEEtran}
\bibliography{references}

\begin{thebibliography}{10}
\providecommand{\url}[1]{#1}
\csname url@samestyle\endcsname
\providecommand{\newblock}{\relax}
\providecommand{\bibinfo}[2]{#2}
\providecommand{\BIBentrySTDinterwordspacing}{\spaceskip=0pt\relax}
\providecommand{\BIBentryALTinterwordstretchfactor}{4}
\providecommand{\BIBentryALTinterwordspacing}{\spaceskip=\fontdimen2\font plus
\BIBentryALTinterwordstretchfactor\fontdimen3\font minus
  \fontdimen4\font\relax}
\providecommand{\BIBforeignlanguage}[2]{{%
\expandafter\ifx\csname l@#1\endcsname\relax
\typeout{** WARNING: IEEEtran.bst: No hyphenation pattern has been}%
\typeout{** loaded for the language `#1'. Using the pattern for}%
\typeout{** the default language instead.}%
\else
\language=\csname l@#1\endcsname
\fi
#2}}
\providecommand{\BIBdecl}{\relax}
\BIBdecl

\bibitem{billard2016learning}
A.~Billard, S.~Calinon, and R.~Dillmann, ``Learning from humans,''
  \emph{Springer Handbook of Robotics, 2nd Ed.}, 2016.

\bibitem{auddy_node}
\BIBentryALTinterwordspacing
S.~Auddy, J.~Hollenstein, M.~Saveriano, A.~Rodríguez-Sánchez, and J.~Piater,
  ``Continual learning from demonstration of robotics skills,'' \emph{Robotics
  and Autonomous Systems}, vol. 165, p. 104427, 2023. [Online]. Available:
  \url{https://www.sciencedirect.com/science/article/pii/S0921889023000660}
\BIBentrySTDinterwordspacing

\bibitem{auddy_snode}
S.~Auddy, J.~Hollenstein, M.~Saveriano, A.~Rodr{\'\i}guez-S{\'a}nchez, and
  J.~Piater, ``Scalable and efficient continual learning from demonstration via
  hypernetwork-generated stable dynamics model,'' \emph{arXiv preprint
  arXiv:2311.03600}, 2023.

\bibitem{kolter2019learning}
J.~Z. Kolter and G.~Manek, ``Learning stable deep dynamics models,''
  \emph{{Advances in Neural Information Processing Systems}}, vol.~32, pp.
  11\,128--11\,136, 2019.

\bibitem{huang2021continual}
Y.~Huang, K.~Xie, H.~Bharadhwaj, and F.~Shkurti, ``Continual model-based
  reinforcement learning with hypernetworks,'' in \emph{2021 IEEE International
  Conference on Robotics and Automation (ICRA)}.\hskip 1em plus 0.5em minus
  0.4em\relax IEEE, 2021, pp. 799--805.

\bibitem{schopf2022hypernetwork}
P.~Sch{\"o}pf, S.~Auddy, J.~Hollenstein, and A.~Rodriguez-Sanchez,
  ``Hypernetwork-ppo for continual reinforcement learning,'' in \emph{Deep
  Reinforcement Learning Workshop NeurIPS 2022}, 2022.

\bibitem{parisi2019CL_survey}
G.~I. Parisi, R.~Kemker, J.~L. Part, C.~Kanan, and S.~Wermter, ``Continual
  lifelong learning with neural networks: A review,'' \emph{Neural Networks},
  vol. 113, pp. 54--71, 2019.

\bibitem{ChangFL20WeightInit}
\BIBentryALTinterwordspacing
O.~Chang, L.~Flokas, and H.~Lipson, ``Principled weight initialization for
  hypernetworks,'' in \emph{8th International Conference on Learning
  Representations, {ICLR} 2020, Addis Ababa, Ethiopia, April 26-30,
  2020}.\hskip 1em plus 0.5em minus 0.4em\relax OpenReview.net, 2020. [Online].
  Available: \url{https://openreview.net/forum?id=H1lma24tPB}
\BIBentrySTDinterwordspacing

\bibitem{he2015delving}
K.~He, X.~Zhang, S.~Ren, and J.~Sun, ``Delving deep into rectifiers: Surpassing
  human-level performance on imagenet classification,'' in \emph{Proceedings of
  the IEEE international conference on computer vision}, 2015, pp. 1026--1034.

\bibitem{urain2020imitationflow}
J.~Urain, M.~Ginesi, D.~Tateo, and J.~Peters, ``Imitationflow: Learning deep
  stable stochastic dynamic systems by normalizing flows,'' in \emph{2020
  IEEE/RSJ International Conference on Intelligent Robots and Systems
  (IROS)}.\hskip 1em plus 0.5em minus 0.4em\relax IEEE, 2020, pp. 5231--5237.

\bibitem{chen2018neural}
R.~T. Chen, Y.~Rubanova, J.~Bettencourt, and D.~Duvenaud, ``Neural ordinary
  differential equations,'' in \emph{Proceedings of the 32nd International
  Conference on Neural Information Processing Systems}, 2018, pp. 6572--6583.

\bibitem{von2019continual}
J.~von Oswald, C.~Henning, J.~Sacramento, and B.~F. Grewe, ``Continual learning
  with hypernetworks,'' in \emph{International Conference on Learning
  Representations ({ICLR})}, 2019.

\bibitem{tieleman20126}
T.~Tieleman and G.~Hinton, ``6.5-rmsprop coursera: Neural networks for machine
  learning university of toronto,'' \emph{Tech. Rep.}, 2012.

\bibitem{kingma2014adam}
D.~P. Kingma and J.~Ba, ``Adam: A method for stochastic optimization,''
  \emph{arXiv preprint arXiv:1412.6980}, 2014.

\bibitem{duchi2011adaptive}
J.~Duchi, E.~Hazan, and Y.~Singer, ``Adaptive subgradient methods for online
  learning and stochastic optimization.'' \emph{Journal of machine learning
  research}, vol.~12, no.~7, 2011.

\bibitem{goodfellow2016deep}
I.~Goodfellow, Y.~Bengio, and A.~Courville, \emph{Deep learning}.\hskip 1em
  plus 0.5em minus 0.4em\relax MIT press, 2016.

\bibitem{glorot2010understanding}
X.~Glorot and Y.~Bengio, ``Understanding the difficulty of training deep
  feedforward neural networks,'' in \emph{Proceedings of the thirteenth
  international conference on artificial intelligence and statistics}.\hskip
  1em plus 0.5em minus 0.4em\relax JMLR Workshop and Conference Proceedings,
  2010, pp. 249--256.

\end{thebibliography}

\end{document}